# Blue and Green-Mode Energy-Efficient Nanoparticle-Based Chemiresistive Sensor Array Realized by Rapid Ensemble Learning

Zeheng Wang,* James Scott Cooper, Muhammad Usman, and Timothy van der Laan

Cite This: https://doi.org/10.1021/acsanm.4c04060

Read Online

ACCESS | Metrics & More | Article Recommendations | Supporting Information

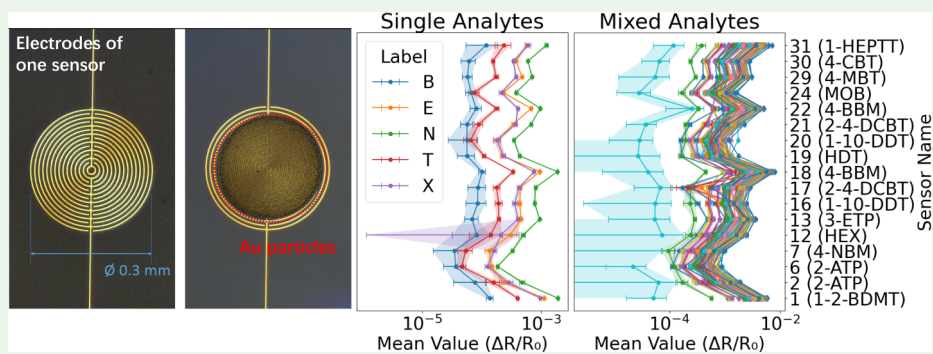

**ABSTRACT:** The rapid advancement of Internet of Things (IoT) necessitates the development of optimized nanoparticle-based chemiresistive sensor (CRS) arrays that are energy-efficient, specific, and sensitive. This study introduces an optimization strategy that employs a rapid ensemble learning-based model committee approach to achieve these goals. Utilizing machine learning models such as Elastic Net Regression, Random Forests, and XGBoost, among others, the strategy identifies the most impactful sensors in a CRS array for accurate classification. A weighted voting mechanism is introduced to aggregate the models' opinions in sensor selection, thereby setting up two distinct working modes, termed "Blue" and "Green". The Blue mode operates with all sensors for maximum detection capability, while the Green mode selectively activates only key sensors, significantly reducing energy consumption without compromising detection accuracy. The strategy is validated through theoretical calculations and Monte Carlo simulations, demonstrating its effectiveness and accuracy. The employed optimization strategy elevates the detection capability of CRS arrays while also pushing it closer to theoretical limits, promising significant implications for the development of low-cost, easily fabricable next-generation IoT sensor terminals.

## 1. INTRODUCTION

Nanoparticle-based chemiresistive sensors (CRSs) are recognized as pivotal sensors within the Internet of Things (IoT)[1−4] network. This is attributed to several key advantages over other sensors, including simple manufacturing,[5] diverse detectable analytes,[6] and rapid readout.[7,8] Over the years, research has demonstrated the versatility of chemiresistive sensors across a multitude of applications. These applications range from medical diagnostics, particularly disease diagnosis,[8] environmental applications such as real-time air quality assessment[9] and the detection of various pollutants in solution.[10,11]

Distinct from traditional analytical instruments, which rely on complex and time-consuming procedures, CRSs operate on a more straightforward principle: the modulation of electrical resistance in the presence of target chemicals.[12,13] This change in resistance is the primary sensing mechanism and is influenced by the material design of the sensor. For instance, metal nanoparticle (MNP)-based sensors and inorganic semiconductor-based sensors each have unique mechanisms that respond to different analytes.[12,14] Despite variations, the core principle remains consistent—analytes induce distinct conductivity changes in the sensors. Thus, by quantitatively measuring resistance fluctuations, the characteristics of the sample under investigation can be inferred.

In practice, CRSs face the challenging task of distinguishing between analytes that possess closely related physiochemical properties—a feat hard to achieve with a single sensor. To work around this sensor arrays or networks are employed.[15−17] Within such an array, each sensor exhibits a unique sensitivity to various analytes, commonly termed "partial selectivity." Leveraging this partial selectivity, the CRS array can function

A

synergistically as a principal component analyzer.[17] In this capacity, it can identify distinct principal components, which may represent specific chemicals or combinations thereof, simply by monitoring the resistance changes across components in the array. Importantly, this approach obviates the need for each individual sensor to exhibit high selectivity toward any specific analyte. Instead, the strength of the array lies in its collective chemical diversity, enabling it to respond to a broader range of analytes with cross-reactive characteristics.[18]

However, the deployment of CRS arrays in modern Internet of Things (IoT) applications presents a significant challenge in terms of energy consumption, as they are inherently resistive. Resistive behavior is naturally energy-intensive necessitating a careful evaluation of each sensor's performance to eliminate redundancy within the array, thereby conserving energy. Similar to other systems that also follow the Net-Zero roadmap,[19,20] this task becomes increasingly complex as the size of the system expands. While various innovative approaches have been proposed to address this issue, e.g., clustering algorithms,[21] search algorithms,[22] and genetic algorithms,[23] these methods often yield sensor subsets optimized for specific analytes, rather than offering a universally applicable solution. Additionally, it has been proved that a single algorithm is typically unsuitable for all circumstances.[24] Therefore, there remains a pressing need for a comprehensive strategy that can effectively utilize less sensors in the array without compromising identification performance across a broad range of analytes.

In this study, we introduce a strategy aimed at optimizing CRS arrays using artificial intelligence (AI)-based ensemble learning techniques. Specifically, we employ a diverse set of eight machine learning algorithms to model the sensor array's behavior. From these, the top-performing models are selected to form a 'model committee.' This committee operates on a weighted voting mechanism to identify essential sensors while eliminating redundant ones, thereby enhancing the array's efficiency. Our approach enables the sensor array to operate in two distinct modes: the "Blue mode," utilizing the full array for maximum sensitivity, and the "Green mode," employing a subset of sensors to achieve comparable sensitivity but with reduced energy consumption. This is important as it would allow the system to generally operate in Green mode to conserve power before switching to Blue mode when required. To validate our strategy, we use the data from experiments comprising of 17 sensors detecting a group of aromatic compounds in a water-based solution.[10] Remarkably, our approach was able to narrow down the active sensors to just 5 in Green mode, with only a 4% reduction in detection capability. We also established a theoretical model that affirms optimization is possible and outlines the theoretical boundary of the optimization. This outcome underscores the potential of our approach for more generalized applications demanding both energy efficiency and high sensitivity, including wearable electronics and smart homes.

## 2. THEORETICAL FRAMEWORK

### 2.1. Theory of Detection Capability of CRS Array.
Before testing this new approach on real data, the theoretical feasibility is established. Constructing a comprehensive model that captures the responses and structures across all possible CRS array configurations is nontrivial and as such the problem is simplified with the following assumptions: 1) the sensors in the array operate independently, 2) their readings never reach saturation, and 3) there is no cross-reactivity among analytes on the sensor surface. Making these assumptions ensures linear contributions to the sensor's resistance change and thus the problem can be reformulated as an optimization problem within the framework of stochastic processes.

In this case, the CRS array's responses to a sample can be depicted by a matrix $D$, where the element $S_{ij}$ represents the i-th sensor's sensitivity to the j-th analyte. Usually, a single sensor in the CRS array will not sense all analytes in the sample effectively, therefore, the matrix $D$ will be sparse, i.e., each line has many zeros indicating that certain sensors cannot respond to certain analytes, such as

$$D_{n \times m} = \begin{pmatrix} S_{11} & 0 & \cdots & S_{1m} \\ 0 & S_{22} & \cdots & \vdots \\ \vdots & S_{ij} & \ddots & 0 \\ S_{n1} & \cdots & 0 & 0 \end{pmatrix} \quad (1)$$

The sample containing a series of analytes can be expressed by another diagonal matrix $X$:

$$X_{m \times m} = \begin{pmatrix} A_{11} & 0 & \cdots & 0 \\ 0 & A_{22} & \cdots & \vdots \\ \vdots & 0 & \ddots & 0 \\ 0 & \cdots & 0 & A_{mm} \end{pmatrix} \quad (2)$$

Therefore, the matrix of one measurement, $M$, can be achieved by using the sensor matrix to multiply the sample matrix, which reads:

$$M = DX = \begin{pmatrix} S_{11}A_{11} & 0 & \cdots & S_{1m}A_{mm} \\ 0 & S_{22}A_{22} & \cdots & \vdots \\ \vdots & 0 & \ddots & 0 \\ S_{n1}A_{11} & \cdots & 0 & 0 \end{pmatrix}_{n \times m} = \begin{pmatrix} R_1 \\ R_2 \\ \vdots \\ R_n \end{pmatrix} \quad (3)$$

Where vector $R_i$ is i-th sensor's readout. Note that the elements of $R_i$ are not explicit in a single readout, as the readout can only produce a single value combining all responses. The tomography of the matrix $M$ can be realized by, e.g., ML algorithms. A perfect measurement requires $M$ to have

$$\exists\, k \leq n:\, \forall\, j \in [1, m], \quad \left(\sum_{i=1}^{k} R_i\right)_j \neq 0 \quad (4)$$

The detection capability of the i-th sensor is quantified by the proportion of nonzero elements in its i-th row. Consequently, the overall detection capability of the CRS array is represented by the proportion of nonzero elements in the vector $\sum_{i=1}^{n} R_i$. For a given analyte j, the sum of the sensors' responses $\left(\sum_{i=1}^{k} R_i\right)_j$ must be nonzero for a single measurement. However, it is important to note that the CRS array may not be fully responsive to all analytes. Some analytes may elicit indistinguishable or even undetectable responses. In such cases, the j-th position in the response vector will be zero, indicating either a lack of response to the j-th analyte or an inability to distinguish this response from those to other analytes.

We assume that each sensor's detection capability (how many analytes it can identify), $C_i$, follows the Gaussian distribution $\mathcal{N}$:

$$1 - \epsilon_i = C_i \sim \mathcal{N}(\mu, \sigma^2) \quad (5)$$

Where $\epsilon$ is the tolerance of the detection capability, which represents the system error rate of one sensor. Then, considering the array's system error $C(n)$ (the percentage of the analytes that the array cannot identify), the optimization problem is (the detailed derivation can be found in Supporting Information):

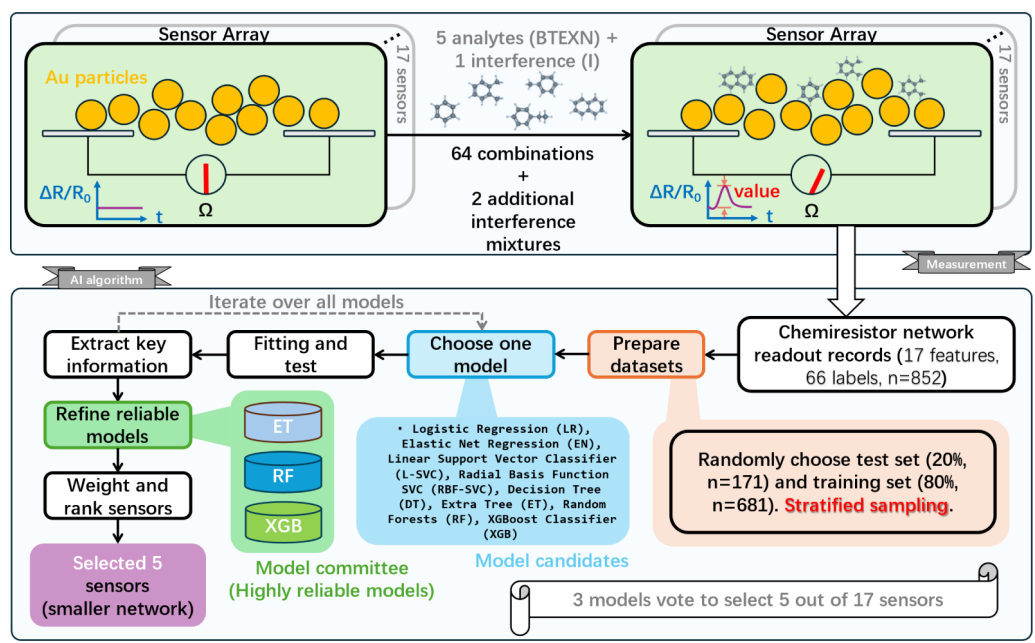

**Figure 1.** Workflow of implementing the proposed strategy for the CRS array data. The measurement and AI process are separated into two panels. The details of the data set and the fabricated sensors can be found in Supporting Information.

**Table 1. Analytes' Contents, Concentrations, and Distributions**[a]

| Analyte | Label | Concentration Range | Distribution |
|---|---|---|---|
| Benzene | B | 44 - 120 µg L$^{-1}$ | |
| Toluene | T | 44 - 110 µg L$^{-1}$ | |
| Ethylbenzene | E | 44 - 120 µg L$^{-1}$ | |
| p-Xylene | X | 44 - 110 µg L$^{-1}$ | |
| Naphthalene | N | 44 - 160 µg L$^{-1}$ | |
| Interferent | I | 62 - 113 µg L$^{-1}$ | |

[a]These ranges are intended to cover variability in real-world scenarios, ensuring that our models are trained and validated across a broad spectrum of possible concentrations.

$$\text{minimize } E[n], \quad s.t. \left[1 - \left(1 - \frac{\mu}{m}\right)^{E[n_{\min}]}\right]^m - C(n) = 0,$$

$$n \in \mathbb{N}, \quad \epsilon \in (0, 1) \tag{6}$$

Note that the whole array's error is the function of the number of the sensor $n$, $C(n)$. Therefore, with the Lagrange multiplier $\lambda$, we will have the Lagrange function:

$$\mathcal{L}(n, \lambda) = E[n] + \lambda \left(C(n) - \left(1 - \left(1 - \frac{\mu}{m}\right)^n\right)^m\right) \tag{7}$$

We can then yield the optimization boundaries of the sensor number of the CRS array (the detailed derivation can be found in Supporting Information):

$$E[n]_{\min} = n \geq \frac{\ln(1 - \sqrt[m]{C})}{\ln\left(1 - \frac{\mu}{m}\right)} \tag{8}$$

Given the number of analytes in the sample, denoted as $m$, and the average detection capability of the CRS array, represented by $\mu$, we can establish a relationship between the minimum required number of sensors, $n$, and the system's detection capability constant, $C$. This demonstrates the feasibility of optimizing the CRS array. However, this theoretical framework only specifies the minimum number of sensors needed for effective measurement, in Green mode; it does not address which sensors should be active. To tackle this remaining question, we will employ our proposed optimization strategy on a CRS array for detecting aromatic compounds in water. Subsequently, we will validate the practical outcomes against the theoretical predictions outlined in eq 8.

**2.2. Optimization Strategy.** The No Free Lunch Theorem in AI research posits that no single AI algorithm can serve as a one-size-fits-all solution for every problem, as evidenced by multiple studies.[24,25] This theorem is particularly relevant when dealing with sensor arrays, which often exhibit diverse structures and behaviors, necessitating the consideration of multiple algorithms for effective optimization.[26,27] Ensemble learning techniques offer a promising avenue to address this, where the final decision will be made by all eligible algorithms together. These techniques have been extensively researched and are known for their ability to mitigate some of the limitations imposed by the No Free Lunch Theorem.[28,29] This work leverages the concept of ensemble learning by incorporating several different algorithms to tackle the challenge of sensor array optimization, as illustrated in Figure 1. The methodology involves the following steps:

● Select explainable model algorithms allowing for the extraction of feature importance.
● Utilize the collected sensing data to train all chosen models and subsequently extract the importance of various features.
● Validate and benchmark the models to identify the best-performing ones.

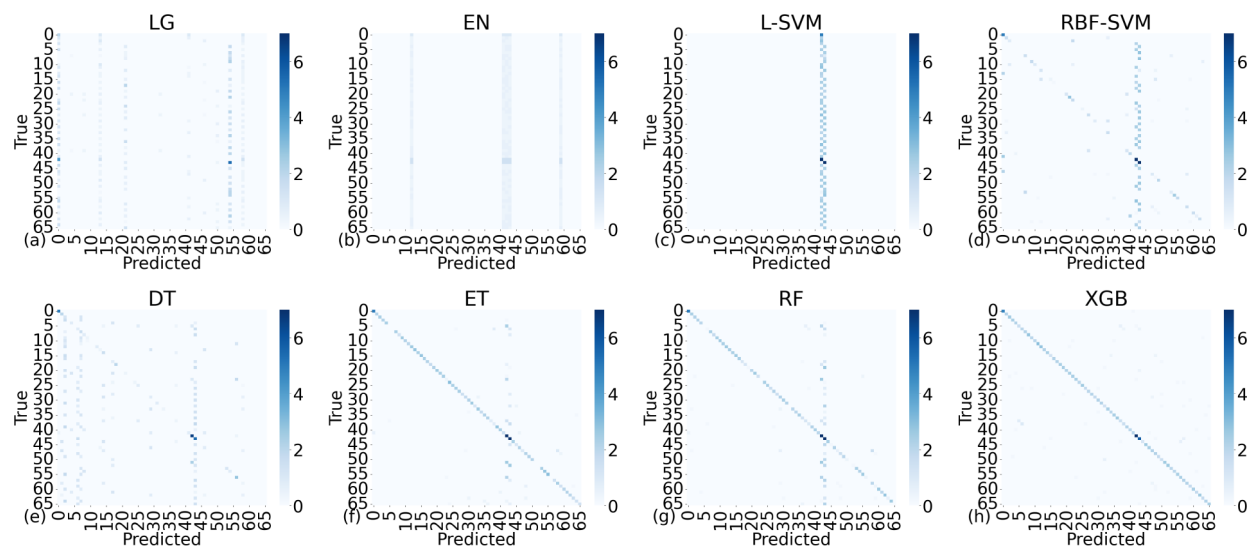

**Figure 2.** Confusion matrices showing the classification performance of the eight machine learning models used for detecting various analytes with the nanoparticle-chemiresistive sensor array. The diagonal elements in each matrix represent the correct classifications, off-diagonal elements indicate misclassifications. The values along the diagonal show the proportion of data points correctly classified for each sample. Higher values on the diagonal and lower off-diagonal values suggest better performance. The Logistic Regression (LG), Elastic Net Regression (EN), and Linear Support Vector Classifier (L-SVC) models (a−c) show limited effectiveness in distinguishing analytes, as indicated by lower values on the diagonal and higher misclassification rates. The Radial Basis Function Support Vector Classifier (RBF-SVC) and Decision Tree (DT) models (d, e) perform moderately, correctly classifying some analytes but struggling with others. The Extra Trees (ET), Random Forest (RF), and XGBoost (XGB) models (f−h) demonstrate strong performance, with high values along the diagonal and minimal off-diagonal errors, indicating a higher overall accuracy in classifying analytes.

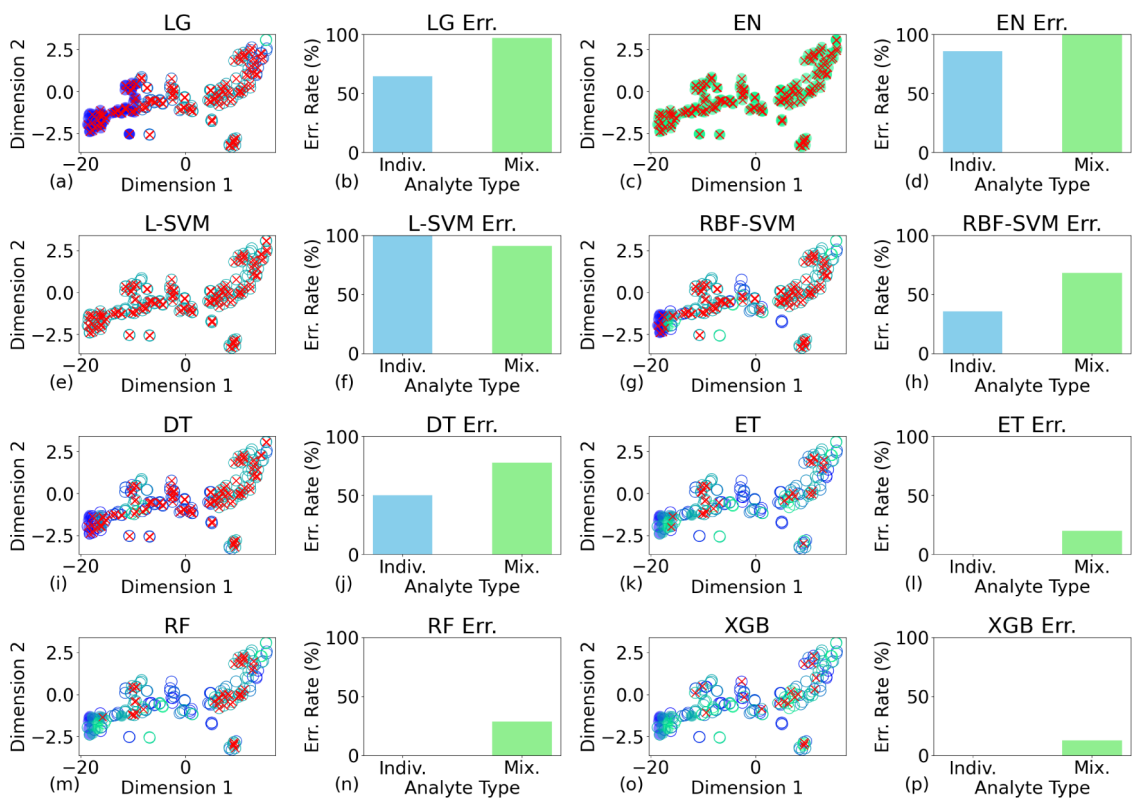

**Figure 3.** First column, (a)−(m), and third column, (c)−(o), the 2D mapping of the data points in one validation shot by tSNE algorithm. Circles with different colors represent different test data, and those with the red error marks are incorrectly classified data points. Second column, (b)−(n), and fourth column, (d)−(p), the corresponding error rates in the classification. These results indicate that ET, RF, XGB are excellent candidates for serving the model committee.

● Employ the selected models in a voting mechanism, weighted by their performance, to choose the most relevant sensors.

● Establish two operational modes for the sensor array: the "Green mode" and the "Blue mode" to fulfill different sensing requirements.

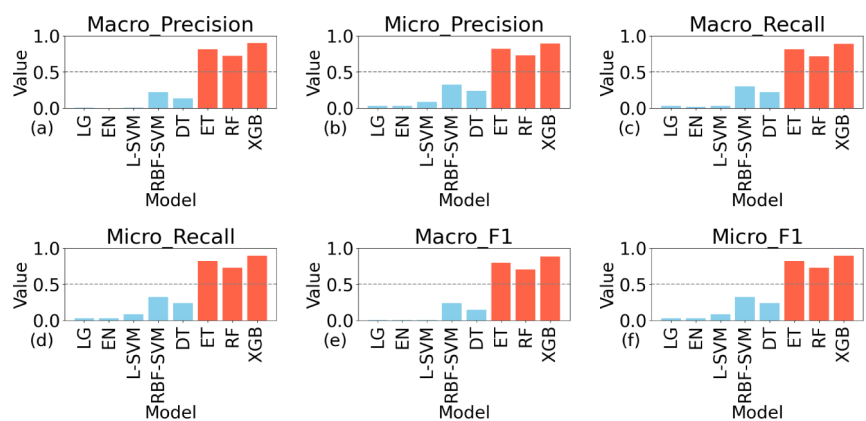

**Figure 4.** Evaluation scores of the classification models. (a)-(f) denote the macro and micro scores, including the precision, recall, and F1, respectively. Only the ET, RF, XGB models (scores highlighted in red) can yield good classifications. Therefore, the ET, RF, and XGB are chosen as the members of the model committee for sensor selection. The models highlighted in red (ET, RF, and XGB) are those chosen to form the final model committee due to their superior performance across all metrics, achieving scores above 0.5, which indicates their higher reliability and accuracy in classifying the analytes. The models not highlighted in red were excluded from the committee due to lower overall performance.

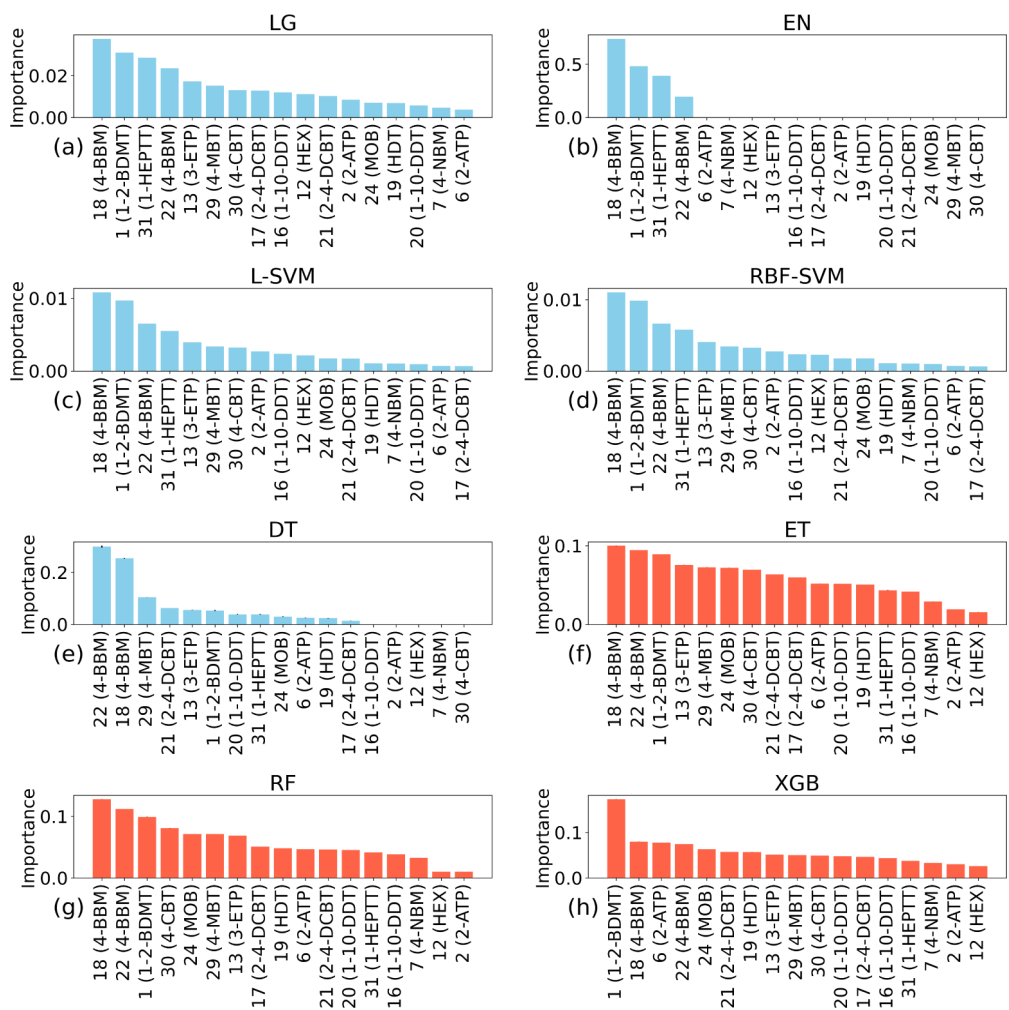

**Figure 5.** Averaged importance ranking of all sensors in the array (calculated by 5 random repeats), listed by eight models. Although other models can extract an importance list (a)−(e), their results were abandoned due to low model performance scores; the high scores for (f) ET, (g) RF, and (h) XGB's, meant these were used in the committee model.

## 3. RESULTS

**3.1. Ensemble Learning Setup.** Figure 1 outlines the workflow for implementing our dual-mode CRS array strategy, utilizing a data set derived from groundwater nanoparticle-CRS data. Details on sensor configurations and data acquisition methodologies can be found elsewhere.[10] In the present study, we employ a data set comprising readout records from 17 sensors, responding to 66 distinct analyte configurations, either individual chemicals or mixtures. This data set encompasses 853 entries, where each entry features 17 sensor readings and a single label identifying the detected analyte. The modeling task is a supervised learning task that aims to train classification models to identify the labels of the CRS array's readout. Comprehensive information on the analytes, including their composition, concentrations, and distribution, is available in Table 1.

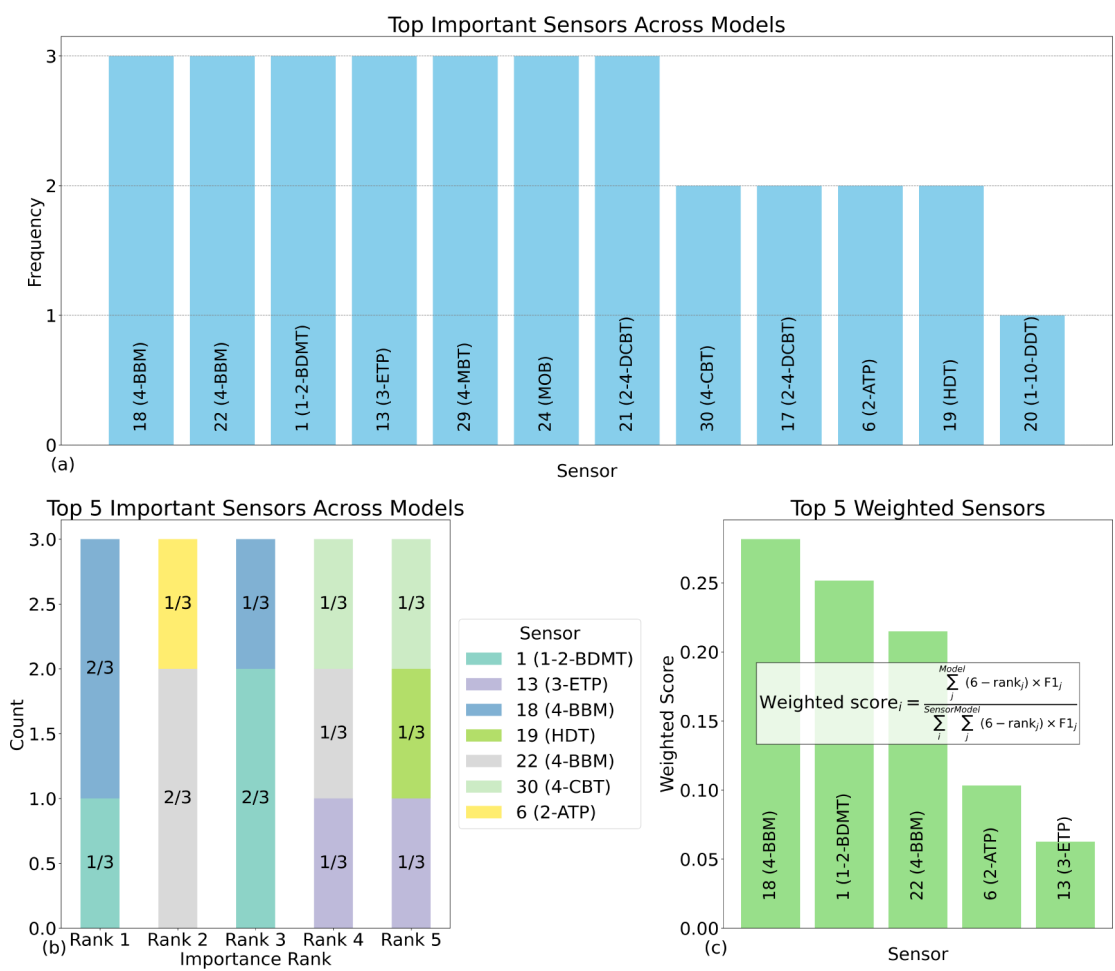

**Figure 6.** Statistics of the sensor selection by the model committee: (a) the frequency, (b) the counts of the ranks, and (c) the weighted scores of the elected sensors across models. Using frequency and ranking counts cannot conclude a straightforward sensor list, while the weighted scores can easily generate a list of selected sensors.

For each training and validation cycle, the data set is partitioned into a training set (comprising 80% of the original experimental data) and a test set (accounting for the remaining 20%). Stratified sampling ensures that both subsets are representative, containing all unique labels. One round of training and validating models and extracting all important features costs only 8.9 s (the details of the hardware specifications can be found in the Supporting Information).

We then employ eight interpretable machine learning models, deliberately excluding deep neural network algorithms, to facilitate rapid and transparent predictive modeling. The selected models include Logistic Regression (LR), Elastic Net Regression (EN), Linear Support Vector Classifier (L-SVC), Radial Basis Function SVC (RBF-SVC), Decision Tree (DT), Extra Tree (ET), Random Forests (RF), and XGBoost Tree Classifier (XGB). Each model is trained using the training set and subsequently validated against the test set. Evaluation metrics are extracted postvalidation, and the entire procedure is detailed in Figure 1. Detailed information of the models can be found in Supporting Information.

**3.2. Model Committee.** To construct the model ensemble, we evaluated the performance of each candidate model. Figure 2 presents the confusion matrices for all eight models, revealing distinct variations in their classification capabilities. Specifically, the ET, RF, and XGB models demonstrate robust performance (refer to the dark line patterns highlighted in the matrices). In the confusion matrices, it is expected that all predictions are correct in the best cases, which will result in a diagonal line. In contrast, LG, EN, and L-SVC exhibit limited efficacy in data modeling post-training (large number of off-diagonal entries). The RBF-SVC and DT models achieve moderate success, as evidenced by the prominence of diagonal elements in their respective confusion matrices.

This performance assessment is further corroborated by a two-dimensional (2D) visualization of classification results (tSNE mapping), depicted in Figure 3. The figure illustrates both the classification outcomes and the corresponding error rates for a single validation run. Red cross markers highlight misclassified data points, providing an intuitive gauge of each model's effectiveness. Notably, the ET, RF, and XGB models excel in classifying individual analytes, with only minor errors observed in the classification of mixed analytes.

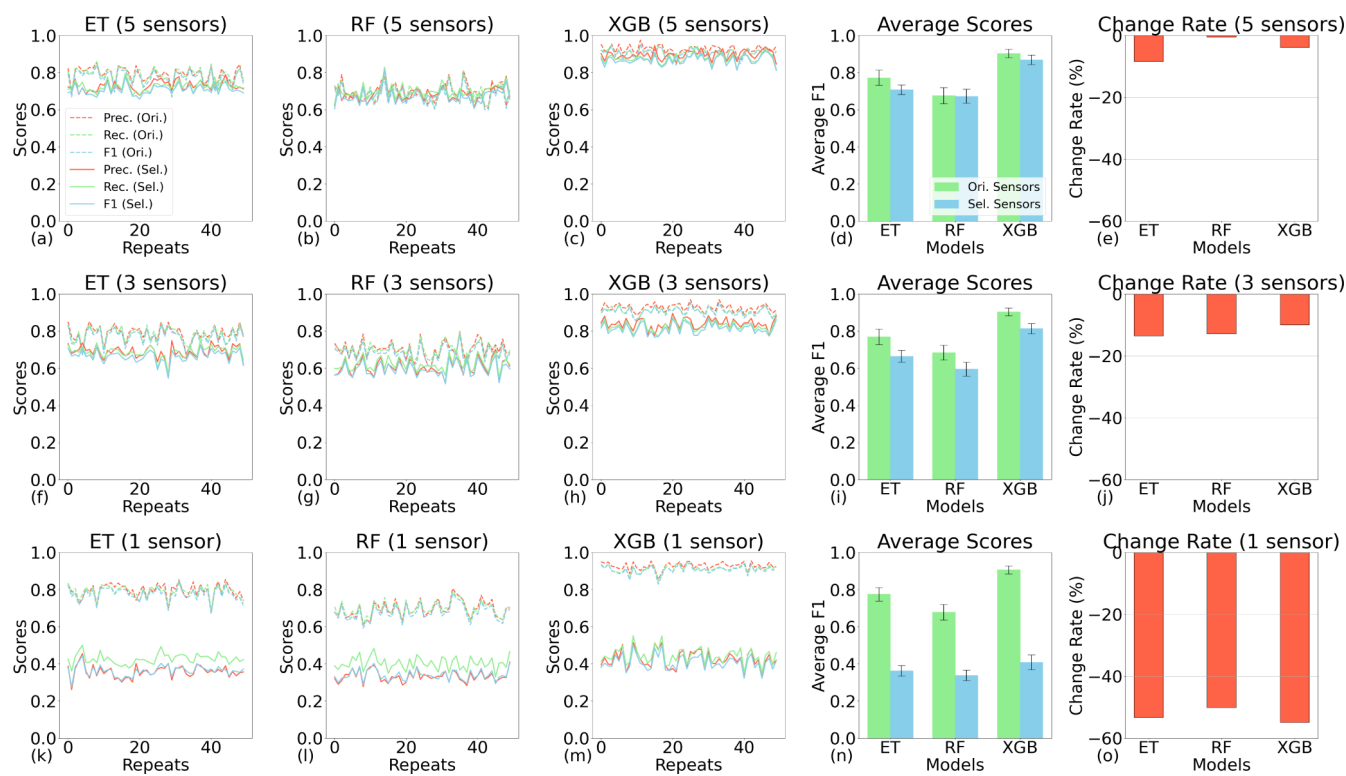

**Figure 7.** Performance scores of the CRS array in different working modes. (a)–(c), (f)–(h), and (k)–(m) are the precision, recall, and F1 scores of the array working in different modes (with different sensors) in multiple validation repeats (each repeat will split the training set and validation set with stratified sampling randomly); (d), (i), and (n) are the average of the F1, with error bars showing the deviation; (e), (j), and (o) are the reduction in average F1 in different working modes (different numbers of activated sensors).

A more objective and comprehensive evaluation is achieved through scoring the candidate models, as depicted in Figure 4. Consistent with the observations from Figures 2 and 3, ET, RF, and XGB emerge as the top-performing models, each achieving scores above 0.7 across all metrics. Notably, XGB appears best with a score of 0.88, affirming its robust modeling capabilities. In contrast, the unhighlighted models fall short, with significantly lower scores and are therefore excluded from the model ensemble. Consequently, the ensemble comprises ET, RF, and XGB.

**3.3. Sensor Array Optimization by Model Committee.** Figure 5 presents the sensor importance rankings for classification across all models. It is important to note, this assessment just considers sensor classification capacity and not sensitivity or energy consumption. We focus on the top three best-performing models, highlighted by red bars, for a more reliable interpretation of sensor importance. Interestingly, the order of sensor importance varies among the considered models.

Relying solely on a single ML model to identify the most impactful sensors for accurate classification is not advisable in this case. Even a straightforward frequency count of sensors appearing in the top 5 positions across each model falls short in pinpointing key performance indicators, as depicted in Figure 6a. A more robust solution is a voting mechanism within the model committee, illustrated in Figure 6b. However, performance disparities exist even among these committee members. Simply averaging their sensor rankings without accounting for these variations could dilute the accuracy of the final selection. To mitigate this, we use the models' F1 scores as weights to fine-tune the ultimate choice of sensors:

$$\text{Weighted score}_i = \frac{\sum_j^{Model} (6 - \text{rank}_i) \times F1_j}{\sum_i^{Sensor} \sum_j^{Model} (6 - \text{rank}_i) \times F1_j} \quad (9)$$

Consequently, the weighted score for the $i$-th sensor is calculated as its rank score $(6 - \text{rank}_i)$ across all models, each weighted by the model's corresponding F1 score $F1_j$. This sum is then normalized by dividing it by the total scores of all sensors across all models in the committee. Utilizing this weighted score, we can compile a list that ranks all sensors based on their impact on sensing, from highest to lowest, as depicted in Figure 6c where we show the top 5 critical sensors.

Leveraging the list of the top 5 critical sensors, we introduce two operational modes for the CRS array: the Blue mode, where all sensors are active for maximum detection capability, and the Green mode, where only the top 5 key sensors are operational to conserve energy with only a small loss of sensing capability. More drastically, we could opt to use only the top 3 most critical sensors for the Green mode, depending on the specific environmental conditions. However, this would reduce sensing capability further as illustrated in Figure 7e,j.

To evaluate the efficacy of the Green mode, we compare the detection capabilities of the nanoparticle-CRS array working in both Blue and Green modes, as illustrated in Figure 7. The first three subfigures in each row of Figure 7 display the array's detection capabilities in varying configurations—using 5, 3, and 1 sensors. Notably, the Green mode with 5 sensors closely matches the Blue mode in detection capability. While fewer sensors result in diminished capability, the 3-sensor Green mode could be advantageous in scenarios where energy conservation is a critical constraint. For this assessment, we employ the validation set and model scores (precision, recall, and F1) as performance metrics.

Figure 7d,i,n offer a direct comparison of F1 scores between the two modes. Utilizing the XGB model for readout, both the 5 and 3-sensor Green modes achieve F1 scores above 0.8, albeit with a slight decline compared to the Blue mode. A more detailed view of this F1 score reduction is presented in Figure 7e,j,o. As show in Table 2 the 5-sensor Green mode

Table 2. Typical Values of Key Evaluation Metrics for Different Machine Learning Algorithms Across Various Operational Modes[a,b]

| Operation Mode | Accuracy | Precision | Recall | F1 | Energy Savings |
|---|---|---|---|---|---|
| Blue Mode | 0.90 | 0.89 | 0.91 | 0.90 | - |
| Green Mode (5 sensors) | 0.86 | 0.84 | 0.85 | 0.84 | 70.6% |
| Green Mode (3 sensors) | 0.80 | 0.77 | 0.78 | 0.77 | 82.4% |

[a]Energy savings are calculated under the assumption that variations in individual sensor power consumption are negligible. [b]More details in Supplementary Information.

significantly reduces energy consumption—by as much as 70% if we assume uniform power budgets across sensors—while only losing less than 5% drop in detection capability. In contrast, the 3-sensor Green mode sees a maximum capability reduction of 15% but offers an even greater 82% reduction in power consumption. Operating with just a single sensor, however, results in a low F1 score, confirming ineffective analyte identification using just a single sensor, which is why arrays are used.

Assuming equal power consumption for each sensor, the energy savings can be calculated as the ratio of the number of active sensors in Green mode to the total number of sensors in Blue mode. For Green mode with 5 sensors, the energy savings = (1−5/17) * 100% ≈ 70.6%. For Green mode with 3 sensors, the energy savings = (1−3/17) * 100% ≈ 82.4%. The extracted key parameters can be found in Table 2.

Thus, by switching from Blue mode to Green mode, we can expect approximately 70−82% energy savings, depending on whether 5 or 3 sensors are active in Green mode. This reduction in energy consumption is significant and demonstrates the effectiveness of the Green mode in conserving power without a substantial loss in detection capability.

**3.4. Validation with Theory.** The efficacy of our optimized strategy is confirmed through both theoretical calculations and Monte Carlo (MC) simulations (500 trials), as depicted in Figure 8. In this figure, the analytical results closely align with the MC simulations, with an error margin of less than one sensor, serving as robust cross-validation. The red stars in the figure represent the outcomes achieved using our proposed Green modes. Usefully the simulation shows that higher individual sensor capability within the nanoparticle-CRS array allows for fewer sensors to achieve a sensing performance as expected. In the experiments used here, average capability across all sensors is 62%, denoted in dark gray. Utilizing the presented committee strategy, we've reduced the sensor count to 5 and 3 in the Green modes (stars in Figure 8). The performance of the green modes both surpass the performance estimates (MC simulations and theoretical calculations) based on a device with 62% capability. This suggests that these sensors perform better than the average. This is indeed the case as the model committee has selected 100%-capability sensors (more details in Supporting Information). This suggests that our strategy adeptly identifies the most effective sensors in the array and leverages their collective, nonlinear working patterns to optimize readout. Consequently, the array's detection capability exceeds the average case and pushes results closer to the theoretical limit.

These findings affirm the promise of our weight-based voting strategy within the model committee for optimizing the CRS array. They also suggest that the energy-efficient Green mode achieved by the proposed strategy could be a steppingstone toward the development of low-cost, easily fabricated next-generation IoT sensors.

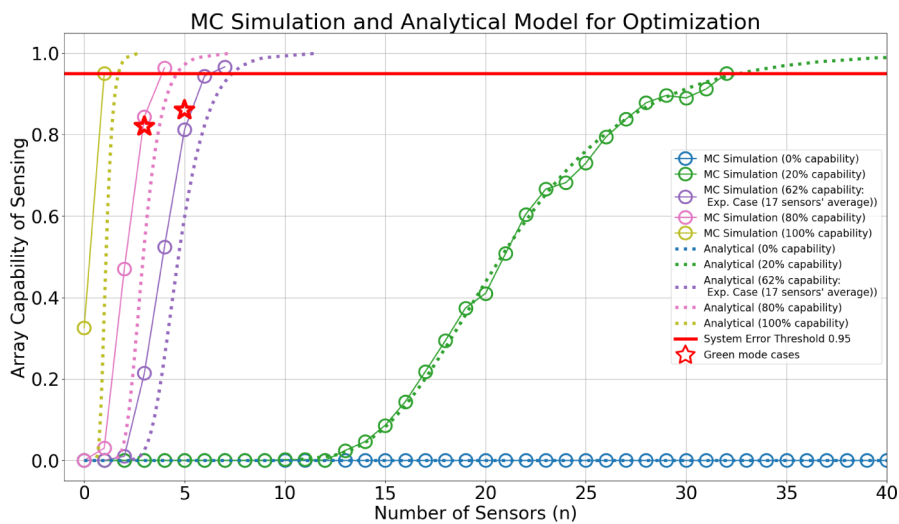

Figure 8. Results of the Green modes benchmarking with the Monte Carlo simulation (500 trials) and the analytical modeling results. The Green mode achieved by the proposed strategy (red stars) can approach the theoretical limit of optimizing the CRS array. The 100% capability's line offsets the vertical limit of 1 due to the variance of the Gaussian distribution in MC simulations.

## 4. CONCLUSION

This study introduces an optimization strategy for nanoparticle-based chemiresistive sensor (CRS) arrays, leveraging a model committee approach to enhance both sensing simplicity and energy efficiency. Validated through theoretical analysis and Monte Carlo simulations, the strategy proves to be both effective and accurate. Specifically, we propose two working modes—Blue and Green—to achieve optimal detection capabilities and energy conservation, respectively.

Our strategy successfully identifies the most impactful sensors for accurate classification and optimizes sensor selection through a weighted voting mechanism within the model committee. This not only elevates the overall detection capability of the array but also significantly reduces energy consumption. The Green mode allows for comparable detection capabilities to the Blue mode, while operating with only five or three key sensors, substantially reducing energy consumption. Furthermore, our approach demonstrates adaptability and scalability, capable of automatically selecting the best sensors within an array and extracting their nonlinear working patterns for optimized readout. This not only enhances the detection capability of the array but also brings it closer to theoretical limits.

In summary, this study presents an effective optimization method for nanoparticle-CRS arrays, promising significant implications for the development of low-cost, easy-to-fabricate next-generation IoT sensors.

## 5. EXPERIMENTAL SECTION

The experimental data for this paper was collected from experiments presented in previous works.[10,11] Briefly, sensors consisted of gold nanoparticle films spread across interdigitated electrodes. The gold nanoparticles of different sensors were functionalized with different thiols that are listed in ref.,[10] The thiols defined the partial selectivity with which different sensors would interact with different analytes, i.e., the 1,10-decanedithiol functionalized sensor (1–10-DDT) had a relatively weak response to naphthalene in comparison to the 1-heptanethiol functionalized sensor (1-HEPTT).

The sensor array was exposed to mixtures of benzene (B), toluene (T), ethylbenzene (E) p-xylene (X), naphthalene (N), and Interferants (I) a "BTEX free" mixture of organics that could potentially interfere with the sensor array's response to B, T, E, X, or N. A full factorial Design of Experiment (DOE) was performed with every possible combination of B, T, E, X, N, I prepared and exposed to the sensor array. Each mixture was exposed to the sensor array 12 times with response (maximum relative resistance change) of every sensor in the array recorded each time. In total the data set contains 66 different mixtures with an occasional repeat giving a total of 71 samples, each exposed 12 times (852 rows) with the concentration of each component (6 columns) and the responses of 17 sensors (17 columns). The original data set is freely available in the supplementary material of ref 10.

## ■ ASSOCIATED CONTENT

### ⓈSupporting Information

The Supporting Information is available free of charge at https://pubs.acs.org/doi/10.1021/acsanm.4c04060.

> Derivation of analytical model for sensor array optimization; description of ensemble learning models and voting mechanism; machine learning model specifications and setup; sensor capability and performance in wet experiments; validation with Monte Carlo simulation; sensor microscopic images (PDF)

## ■ AUTHOR INFORMATION

**Corresponding Author**

**Zeheng Wang** − *Data61, CSIRO, Clayton, Victoria 3168, Australia; Manufacturing, CSIRO, West Lindfield, New South Wales 2070, Australia;* orcid.org/0000-0002-6994-1234; Email: zenwang@outlook.com

**Authors**

**James Scott Cooper** − *Manufacturing, CSIRO, West Lindfield, New South Wales 2070, Australia*

**Muhammad Usman** − *Data61, CSIRO, Clayton, Victoria 3168, Australia; School of Physics, The University of Melbourne, Parkville, Victoria 3010, Australia*

**Timothy van der Laan** − *Manufacturing, CSIRO, West Lindfield, New South Wales 2070, Australia*

Complete contact information is available at:
https://pubs.acs.org/10.1021/acsanm.4c04060

**Notes**

The authors declare no competing financial interest.

## ■ ACKNOWLEDGMENTS

This work was partially supported by CSIRO Impossible-Without-You program.